\documentclass[letterpaper, 10 pt, conference]{IEEEtran}
\IEEEoverridecommandlockouts

\usepackage{cite}
\usepackage{amsmath,amssymb,amsfonts}
\usepackage{algorithmic}
\usepackage{graphicx}
\usepackage{textcomp}
\usepackage{multirow}
\usepackage[font=small]{caption}
\usepackage{subcaption}
\usepackage[table,xcdraw]{xcolor}
\usepackage{hyperref}
\usepackage{balance}

\def\BibTeX{{\rm B\kern-.05em{\sc i\kern-.025em b}\kern-.08em
    T\kern-.1667em\lower.7ex\hbox{E}\kern-.125emX}}

\makeatletter
    \newcommand{\linebreakand}{%
      \end{@IEEEauthorhalign}
      \hfill\mbox{}\par
      \mbox{}\hfill\begin{@IEEEauthorhalign}
    }
\makeatother

\begin{document}

\title{GazeGrasp: DNN-Driven Robotic Grasping with Wearable Eye-Gaze Interface}
% \author {The list of authors has been anonymized for submission
% \thanks{Anonymized for submission}}
% \author{Issatay Tokmurziyev, Miguel Altamirano Cabrera, Luis Moreno, Muhammad Haris Khan, Dzmitry Tsetserukou\\

\author{\IEEEauthorblockN{Issatay Tokmurziyev}
\IEEEauthorblockA{\textit{Skoltech} \\
Moscow, Russia \\
Issatay.Tokmurziyev@skoltech.ru}
\and
\IEEEauthorblockN{Miguel Altamirano Cabrera}
\IEEEauthorblockA{\textit{Skoltech} \\
Moscow, Russia \\
m.altamirano@skoltech.ru}
\and
\IEEEauthorblockN{Luis Moreno}
\IEEEauthorblockA{\textit{Skoltech} \\
Moscow, Russia \\
Luis.Moreno@skoltech.ru}
\linebreakand
\IEEEauthorblockN{Muhammad Haris Khan}
\IEEEauthorblockA{\textit{Skoltech} \\
Moscow, Russia \\
Haris.Khan@skoltech.ru}
\and
\IEEEauthorblockN{Dzmitry Tsetserukou}
\IEEEauthorblockA{\textit{Skoltech} \\
Moscow, Russia \\
d.tsetserukou@skoltech.ru}}

% \thanks{The authors are with the Intelligent Space Robotics Laboratory, Center for Digital Engineering, Skolkovo Institute of Science and Technology. 
% {\tt \{issatay.tokmurziyev, m.altamirano, luis.moreno, haris.khan,  d.tsetserukou\}@skoltech.ru}}
% }

\maketitle

\begin{abstract}
We present GazeGrasp, a gaze-based manipulation system enabling individuals with motor impairments to control collaborative robots using eye-gaze. The system employs an ESP32 CAM for eye tracking, MediaPipe for gaze detection, and YOLOv8 for object localization, integrated with a Universal Robot UR10 for manipulation tasks. After user-specific calibration, the system allows intuitive object selection with a magnetic snapping effect and robot control via eye gestures. Experimental evaluation involving 13 participants demonstrated that the magnetic snapping effect significantly reduced gaze alignment time, improving task efficiency by 31\%. GazeGrasp provides a robust, hands-free interface for assistive robotics, enhancing accessibility and autonomy for users.
\end{abstract}

\begin{IEEEkeywords}
\textit{Eye-Gaze Tracking; Human-Robot Interaction; Assistive Robotics}
\end{IEEEkeywords}
% Add CCS concepts (manually display as plain text in IEEE)
% \textbf{CCS Concepts:}

% $\bullet$ \textbf{Human-centered computing} $\rightarrow$ \textit{Collaborative interaction} $\bullet$ \textbf{Computing methodologies} $\rightarrow$ \textit{Tracking} $\bullet$ \textbf{Social and professional topics} $\rightarrow$ \textit{Assistive technologies}

% \begin{CCSXML}
% <ccs2012>
%    <concept>
%        <concept_id>10003120.10003121.10003124.10011751</concept_id>
%        <concept_desc>Human-centered computing~Collaborative interaction</concept_desc>
%        <concept_significance>100</concept_significance>
%        </concept>
%    <concept>
%        <concept_id>10010147.10010178.10010224.10010245.10010253</concept_id>
%        <concept_desc>Computing methodologies~Tracking</concept_desc>
%        <concept_significance>300</concept_significance>
%        </concept>
%    <concept>
%        <concept_id>10003456.10003457.10003580.10003587</concept_id>
%        <concept_desc>Social and professional topics~Assistive technologies</concept_desc>
%        <concept_significance>300</concept_significance>
%        </concept>
%  </ccs2012>
% \end{CCSXML}

% \ccsdesc[100]{Human-centered computing~Collaborative interaction}
% \ccsdesc[300]{Computing methodologies~Tracking}
% \ccsdesc[300]{Social and professional topics~Assistive technologies}

\section{Introduction}
In recent years, there has been a growing interest in using novel human-machine interfaces to help people with disabilities interact with their surroundings. Gaze-based control systems are promising because they offer a natural and intuitive way to communicate and control devices. Eye-gaze control enables users to interact with robotic systems by simply moving their eyes, which can be especially beneficial for individuals with motor impairments who might find it challenging to use manual input devices. For instance, research has demonstrated the effectiveness of eye-gaze control in operating assistive robotic arms for activities of daily living~\cite{ieee_assistive_device}.

\begin{figure}[t]
\centerline{\includegraphics[width=0.5\textwidth]{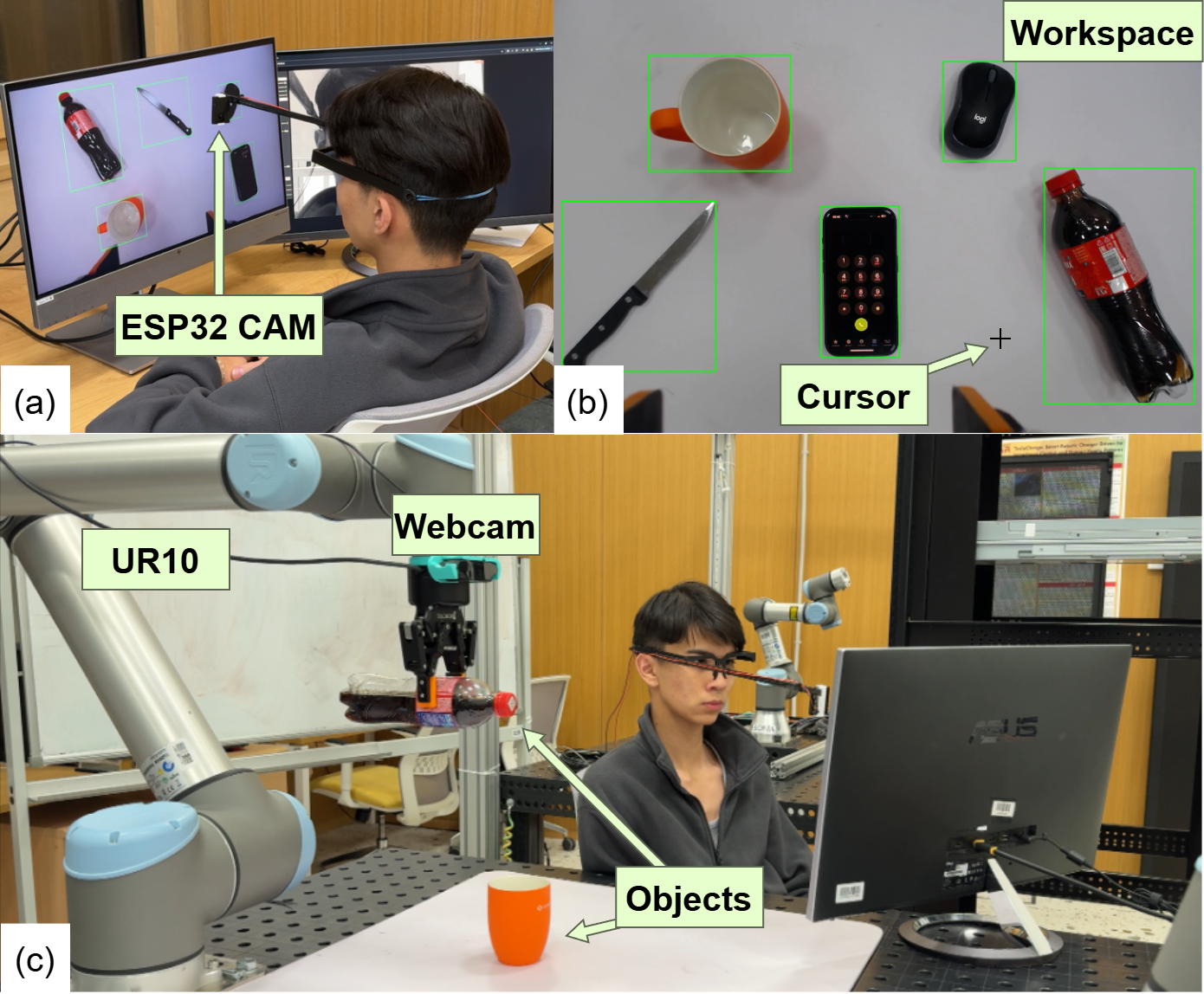}}
\caption{GazeGrasp interface for controlling the collaborative robot UR10 using a gaze-based tracking system: (a) The calibration stage for the users. (b) The workspace view with highlighted objects. (c) Experimental setup for the evaluation of gaze-based control.}
\label{fig:first}
\vspace{-0.6cm}
\end{figure}

This paper introduces \textbf{GazeGrasp}, a novel system that harnesses eye-gaze control for the manipulation of objects via a collaborative robot interface. The system is designed to enable disabled users to control robotic platforms, such as collaborative robots or drones, using only their gaze. Our implementation features a Universal Robot 10 (UR10) collaborative robot, allowing users to perform essential tasks like object grasping and placement. The system interface is illustrated in Fig.~\ref{fig:first}. GazeGrasp employs in-house eye-tracking technologies to accurately interpret gaze direction and convert it into precise robotic commands, delivering a fully hands-free and accessible interface for individuals with severe motor impairments.

The potential of gaze-based manipulation extends beyond accessibility, with applications in healthcare, rehabilitation, logistics, and industrial automation. By enabling natural and intuitive control, these systems enhance human-robot interaction, reduce cognitive load, and improve the usability of robotic platforms. Recent advancements in gaze-based intention recognition~\cite{acm_transparency} and augmented reality integration~\cite{acm_decision_framework} have further demonstrated the versatility and scalability of these interfaces in both assistive and industrial domains.

% In this work, we present the design, development, and experimental evaluation of GazeGrasp. The system demonstrates the feasibility of gaze-based manipulation for collaborative robots and explores its broader implications in assistive technology. Through this research, we aim to advance the state of the art in human-robot interaction and contribute to developing inclusive robotic solutions for individuals with physical impairments.

\section{Related Work}

Eye-gaze control has become an essential modality in advancing human-robot interaction, offering intuitive and accessible interfaces for a variety of applications. From assistive robotics to collaborative systems and teleoperation, recent developments highlight the transformative potential of gaze-based systems~\cite{gazeintention}.

\subsection{Gaze-Based Robotic Control}

Eye-gaze control has proven effective across a wide range of robotic systems, facilitating tasks such as navigation and manipulation. Yuan et al.~\cite{EyeMAV} demonstrated a gaze-driven spatial tasking system for micro aerial vehicles (MAVs), enabling intuitive navigation in three-dimensional environments by decoupling gaze direction from head orientation. Building on this, Wang et al.~\cite{GPA_Teleoperation} introduced a perception-aware teleoperation framework that uses gaze to capture operator intent and generate safe trajectories, making robotic control accessible to non-experts.

In assistive robotics, systems like GazeRace translate gaze movements into precise navigational commands, minimizing cognitive load while enhancing operational accuracy~\cite{GazeRace1}. Recent approaches further refine precision by integrating fiducial markers with gaze control, showcasing the potential for high-accuracy manipulation tasks~\cite{Diegetic2024}. These advancements underscore the role of gaze as a primary input modality for robotic systems, particularly for users with limited motor abilities.

\subsection{Collaborative Robot Interfaces}

Collaborative robots, or cobots, have increasingly adopted intuitive input modalities to enhance interaction in shared workspaces. Sautenkov et al.~\cite{CobotTouch} developed an augmented reality-based system for controlling the UR10 cobot, combining tactile feedback with visual interfaces to improve precision and usability. Fortini et al.~\cite{UpperLimb} explored gaze-based control for upper-limb assistance using the Franka Emika Panda robotic arm, demonstrating the feasibility of gaze-driven task execution in rehabilitation scenarios.

The integration of augmented reality with gaze control has shown promise in advancing collaborative robotics. A recent study highlights the efficiency gains achieved by leveraging eye-gaze in hands-free operations, further expanding the applicability of cobots in industrial and assistive domains~\cite{AugmentedReality2023}. These efforts illustrate the potential of gaze-based interfaces to streamline cobot operation while maintaining safety and efficiency.

\subsection{Remote Manipulation Systems}

Teleoperation interfaces have evolved to bridge the gap between human intent and robotic execution through innovative input mechanisms. Systems such as OmniCharger and OmniRace have introduced gesture-based controls for robotic teleoperation in specialized contexts, such as drone racing and teleconferencing~\cite{OmniCharger, OmniRace}. However, these systems often rely on auxiliary hardware, limiting their accessibility for users with severe motor impairments.

Eye-gaze control offers a more inclusive alternative, as demonstrated by recent work on gaze-controlled telepresence robots~\cite{Telepresence2019}. These systems enable individuals with motor impairments to interact in professional and social settings, although their focus remains on remote presence rather than direct physical manipulation. This highlights a gap in leveraging gaze control for manipulation tasks involving collaborative robots.

Existing gaze-based systems often lack a unified framework for remote manipulation, particularly in collaborative robotic environments~\cite{gazeautonomy}. GazeGrasp addresses this limitation by providing a robust system for object manipulation using only eye gaze as input. The use of gaze as the sole control modality eliminates the need for auxiliary devices, simplifying the interaction process while maintaining high precision.

\begin{figure}[t]
\centerline{\includegraphics[width=0.5\textwidth]{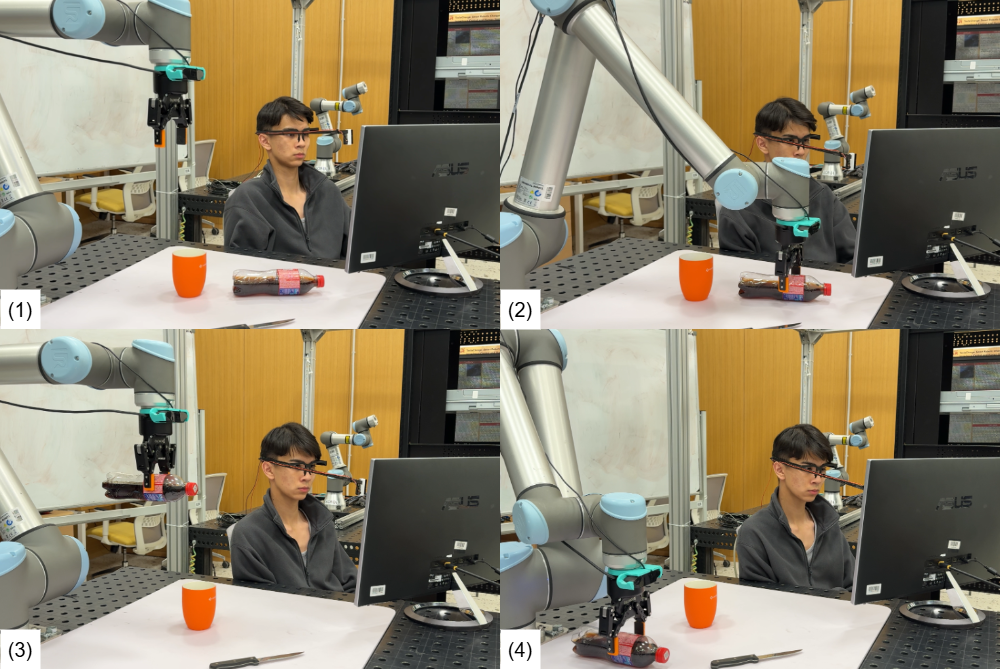}}
\caption{The flow of the task execution: (1) User aims at the object. (2) UR10 goes to the exact position and picks the object. (3) UR10 returns to the initial position. (4) User chooses a new position to put the object.}
\label{fig:ur10}
\vspace{-0.4cm}
\end{figure}

\section{System Overview}

\begin{figure*}[t]
    \centering
    \includegraphics[width=0.8\textwidth]{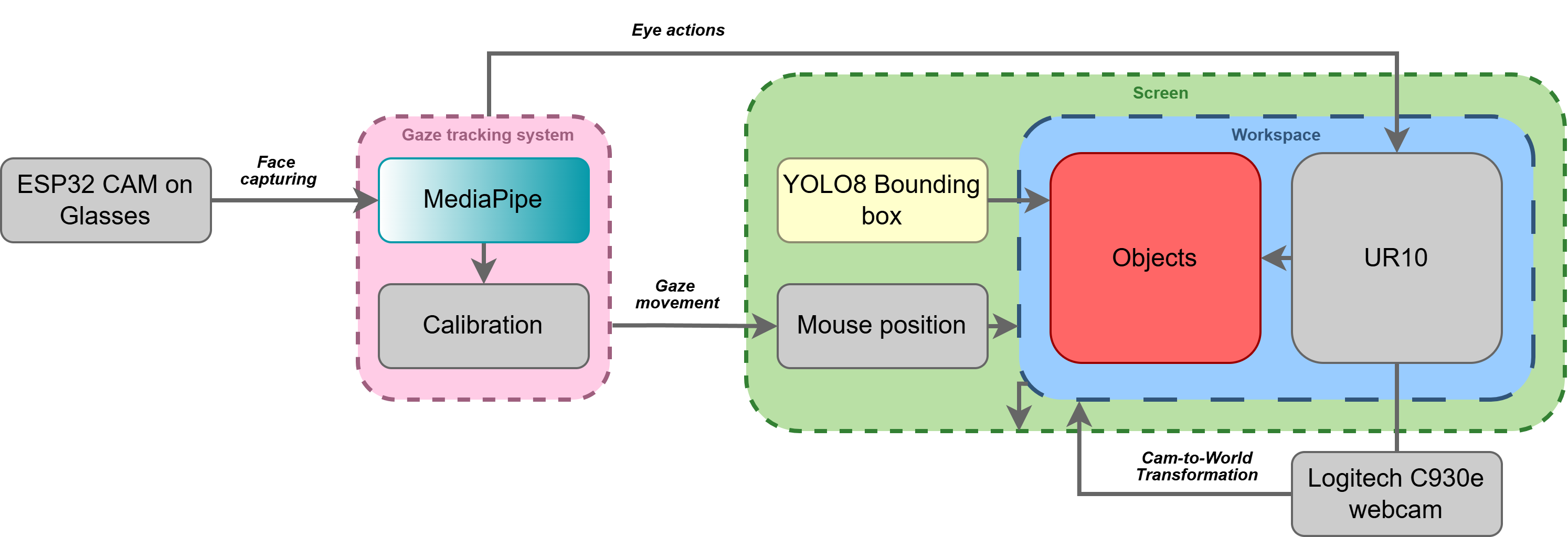} 
    \caption{System Architecture of GazeGrasp: The system integrates gaze-tracking, object detection, and robotic control to enable intuitive manipulation through gaze-based interaction.}
    \label{fig:system_architecture}
\end{figure*}

\begin{figure}[t]
\centerline{\includegraphics[width=0.5\textwidth]{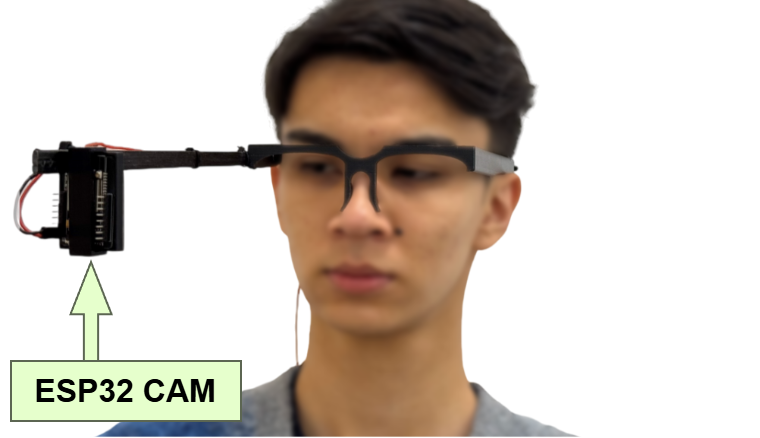}}
\caption{Glasses equipped with an ESP32 CAM. The video is transmitted from the server, allowing remote control.}
\label{fig:cam}
\vspace{-0.4cm}
\end{figure}

At a high level, GazeGrasp works as follows: wearing glasses equipped with an ESP32 CAM (see Fig.~\ref{fig:cam}), the user performs a calibration for their gaze so the system knows where they are looking on the screen. Then, the user sees a live stream from a Logitech C930e webcam located on the end-effector of the UR10 robot. The workspace contains several objects (cup, knife, bottle, phone, mouse), each detected by YOLOv8 and highlighted with bounding boxes. By simply looking at an object for 3 seconds, the user commands the UR10 to pick it up. To help with precision, when the user's gaze enters an object's bounding box, the gaze-controlled cursor snaps (magnets) to the center of that box. The Robotiq 2F-85 gripper grabs the objects. After picking up the object, the user can look at an empty spot in the workspace for 3 seconds, prompting the robot to place the object there. Fig.~\ref{fig:ur10} shows the flow of the task execution with picking and placing the objects. All actions—selection, grasping, placing—are done using eye gaze alone, making the system intuitive and hands-free.

A key component of the gaze detection pipeline is the use of MediaPipe, a framework that robustly detects and tracks facial landmarks. MediaPipe leverages deep neural network (DNN) models to detect and track facial landmarks, enabling precise iris localization and real-time gaze direction estimation critical for the GazeGrasp system~\cite{MediapipePaper}. This information is then mapped onto the screen coordinates, ensuring that the gaze-controlled cursor movements are both stable and responsive. MediaPipe’s reliability in extracting precise eye-related landmarks greatly contributes to the accuracy and smoothness of the GazeGrasp interface, even under varying lighting conditions. The entire system architecture is depicted in Fig.~\ref{fig:system_architecture}.

We now provide a detailed technical description of the system, including how gaze is calibrated, how noise is filtered, how bounding boxes guide object selection, and how pixel coordinates are translated into real-world robot coordinates using camera-to-base transformations.

\subsection{Calibration via Polynomial Regression}

The calibration begins with the user focusing on $N=35$ predefined points displayed on the screen. The system maps iris coordinates $(u, v)$ from the ESP32 CAM to the screen coordinates $(x_{\text{screen}}, y_{\text{screen}})$ using a polynomial regression model:

\begin{equation}
\begin{aligned}
x_{\text{screen}} &= f(u, v) = \sum_{i=0}^{d}\sum_{j=0}^{d-i} a_{ij} u^i v^j, \\
y_{\text{screen}} &= g(u, v) = \sum_{i=0}^{d}\sum_{j=0}^{d-i} b_{ij} u^i v^j,
\end{aligned}
\end{equation}
where $d=3$ is the polynomial degree, and $a_{ij}$ and $b_{ij}$ are coefficients learned during calibration. The calibration ensures that the system accurately maps eye movements to screen coordinates, creating a reliable foundation for gaze-based control. By focusing on a series of screen locations, the user provides the data needed to calculate these coefficients, aligning the system’s gaze estimates with the physical screen layout.

\subsection{Real-Time Gaze Tracking with Kalman Filters}

To stabilize gaze tracking during operation, the system applies a Kalman filter, which smooths the raw iris coordinate data. The filter operates as follows:

\begin{equation}
\mathbf{x}_k = \mathbf{A}\mathbf{x}_{k-1} + \mathbf{w}_k,\quad
\mathbf{z}_k = \mathbf{H}\mathbf{x}_k + \mathbf{v}_k,
\end{equation}
where $\mathbf{x}_k$ is the estimated state (smoothed gaze), $\mathbf{z}_k$ is the observed state (raw gaze), and $\mathbf{w}_k, \mathbf{v}_k$ are process and observation noise, respectively. The Kalman filter reduces noise in the gaze data, ensuring smooth and accurate screen cursor movement even in dynamic conditions.

\subsection{Object Detection and Magnetic Snapping}

Object detection is handled by YOLOv8, which identifies objects within the workspace and outputs bounding boxes:

\begin{equation}
\text{bbox} = \{x_c, y_c, w, h\},
\end{equation}
where $(x_c, y_c)$ is the center of the bounding box and $w, h$ are its width and height. When the user’s gaze enters an object’s bounding box, the system activates a "magnetic snapping" effect:

\begin{equation}
\mathbf{p}_{\text{mouse}} =
\begin{cases}
(x_c, y_c), & \text{if inside\ bbox},\\
\mathbf{p}_{\text{gaze}}, & \text{otherwise}.
\end{cases}
\end{equation}

This feature simplifies precise object selection by automatically snapping the cursor to the center of the object, reducing the need for fine gaze adjustments and improving the overall user experience.

\subsection{Transformation from Camera to Robot Base Coordinates}

For the robot to manipulate objects, the detected bounding box center $(x_c, y_c)$ must be transformed into real-world coordinates $(X, Y, Z)$ in the robot's workspace. This transformation relies on the camera's intrinsic and extrinsic parameters:

\begin{equation}
\begin{bmatrix} X \\ Y \\ Z \end{bmatrix} 
= \mathbf{T}_{\text{base}}^{\text{camera}} \cdot \Pi^{-1}(x_c, y_c),
\end{equation}
where $\Pi^{-1}$ is the inverse projection function and $\mathbf{T}_{\text{base}}^{\text{camera}}$ is a known transformation matrix linking the camera frame to the robot’s base frame. For planar workspaces, the relationship simplifies to:

\begin{equation}
\begin{bmatrix} X \\ Y \\ Z \end{bmatrix}
= \mathbf{H}
\begin{bmatrix} x_c \\ y_c \\ 1 \end{bmatrix}
\end{equation}

This mapping ensures that the robot can accurately position its end effector relative to the detected object, enabling seamless object manipulation.

\subsection{Picking and Placing Objects}

After determining the object’s real-world coordinates, the robot executes the pick-and-place operation. The user initiates these actions by maintaining gaze on the object for three seconds. The robot moves its gripper to the location $(X, Y, Z)$ to pick up the object and subsequently places it at $(X_{\text{empty}}, Y_{\text{empty}}, Z_{\text{empty}})$ based on gaze input. This process is entirely hands-free, leveraging gaze as the sole input modality to complete the manipulation tasks effectively and intuitively.

\section{Experimental Evaluation}

In preliminary testing, we evaluated the effect of the magnetic snapping feature on task completion time. We conducted a study involving 13 participants who agreed to participate in the evaluation, comprising five women and eight men aged between 21 and 37. Their task was to evaluate the functionality of our gaze-based system. Participants were instructed to look at specific objects on the screen, both with and without the magnetic effect of the cursor. They were presented with randomly ordered objects and required to fixate on them 40 times. Initially, they focused on the objects without the magnetic effect of the cursor, followed by the presence of the magnetic effect. The time taken for each participant to successfully fixate on the objects was recorded. Each participant underwent a training session consisting of the task description and calibration phase to ensure accurate data collection.

We investigated the accuracy of gaze-based object selection, particularly comparing scenarios with and without the magnetic effect. We hypothesize that the magnetic effect will significantly enhance system precision by reducing the effort required to align the cursor with the object centers.

\begin{figure}[t]
\centerline{\includegraphics[width=0.4\textwidth]{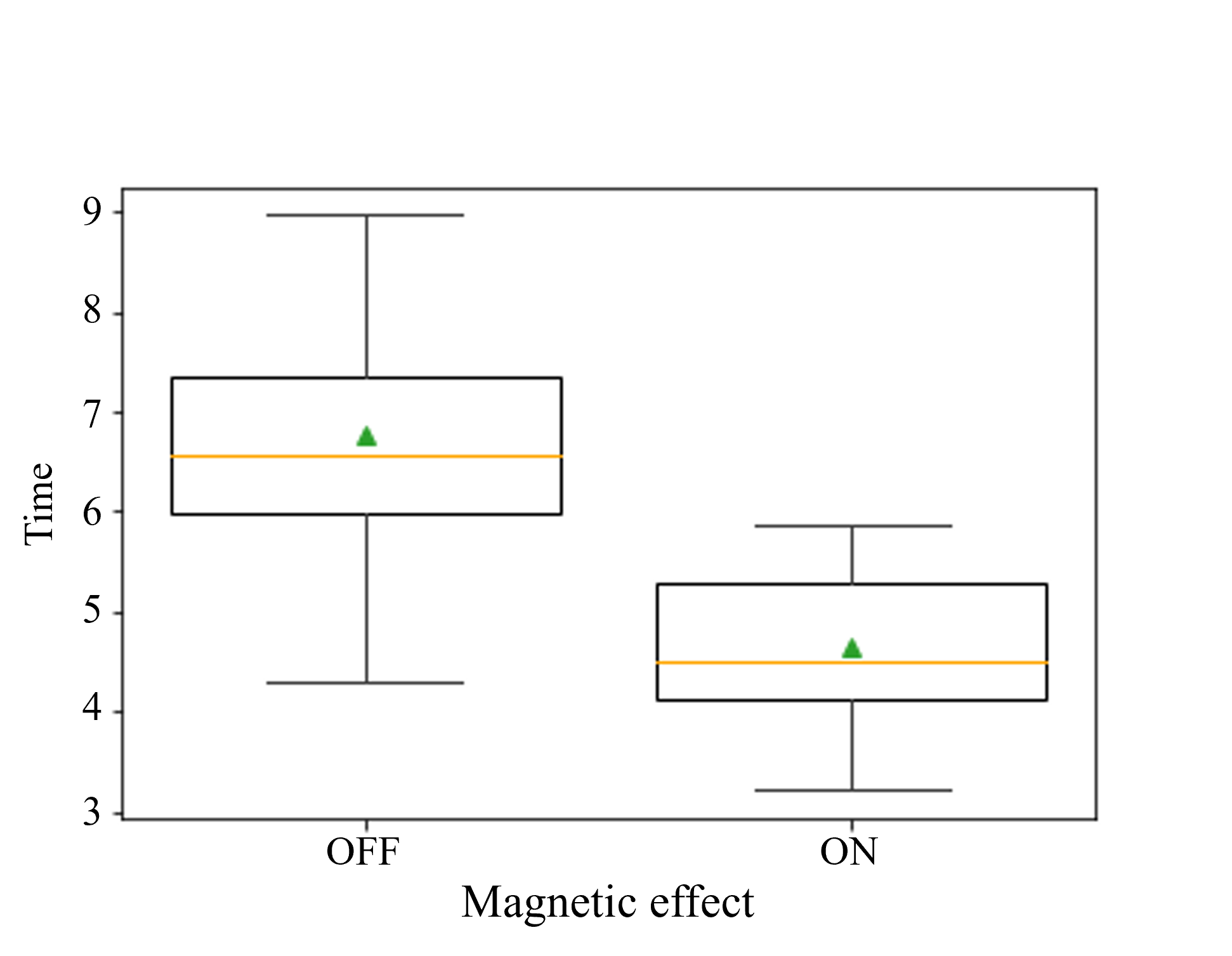}}
\caption{Box plot showing the time required for users to align their gaze with the center of the object under two conditions: with magnetic assistance (ON) and without magnetic assistance (OFF).}
\label{fig:times}
\vspace{-0.4cm}
\end{figure}

The results show that the average time without magnetic effect was 6.77 sec and with magnetic effect 4.65 sec. In order to evaluate the statistical significance of the differences between the modalities from each of the users, we analyzed the results using a single-factor repeated-measures ANOVA with a chosen significance level of $\alpha<0.05$. The open-source statistical packages Pingouin and Stats models were used for the statistical analysis. The results are visualized in Fig.~\ref{fig:times}.

According to the ANOVA results, there is a statistically significant difference in the time of the users while using the two different modalities: $F(1,24) =  24.5204$, $p = 0.000047$. The ANOVA showed that the implementation of the magnetic effect during the manipulation of the objects significantly influenced the completion time of the task.

\section{Conclusions and Future Work}

We introduced GazeGrasp, an innovative gaze-based manipulation system enabling intuitive, hands-free robotic control for individuals with motor impairments. Through robust integration of MediaPipe for gaze detection, YOLOv8 for object localization, and precise robotic control using the Universal Robot UR10, we demonstrated the feasibility of manipulation tasks driven solely by eye movements.

Experimental results from 13 participants showed that the magnetic snapping effect significantly improved task efficiency, reducing average gaze alignment time from 6.77 sec (without snapping) to 4.65 sec (with snapping). Statistical analysis confirmed that the difference was significant ($F(1, 24) = 24.52, p < 0.001$).

% This work emphasizes the potential of gaze-based systems in advancing assistive robotics by reducing the reliance on manual input devices. GazeGrasp provides an accessible and efficient solution for individuals with severe motor impairments, enhancing their autonomy in interacting with their environment.

Future directions include conducting extensive user studies to assess the system's long-term usability, optimizing camera-to-robot calibration for enhanced spatial accuracy, and incorporating the necessary logic for operating the system in cluttered environments, including obstacle avoidance. With continued development, GazeGrasp could play a pivotal role in shaping inclusive robotic systems for rehabilitation, healthcare, and assistive technologies.

\section*{Acknowledgements} 
Research reported in this publication was financially supported by the RSF grant No. 24-41-02039.

\bibliographystyle{IEEEtran}
\bibliography{ref}
\balance

\end{document}